\documentclass[10pt,twocolumn,letterpaper]{article}

\usepackage{iccv}              

\usepackage{bbding}

\definecolor{iccvblue}{rgb}{0.21,0.49,0.74}
\usepackage[pagebackref,breaklinks,colorlinks,allcolors=iccvblue]{hyperref}

\def\paperID{9206} 
\def\confName{ICCV}
\def\confYear{2025}

\title{NavQ: Learning a Q-Model for Foresighted Vision-and-Language Navigation}

\author{
Peiran Xu\quad
Xicheng Gong\quad
Yadong Mu\thanks{Corresponding author.}\\
Peking University\\
Beijing, China \\
{\tt\small xpr820@pku.edu.cn\quad gongxicheng@stu.pku.edu.cn\quad myd@pku.edu.cn}
}

\begin{document}
\maketitle

\begin{abstract}
In this work we concentrate on the task of goal-oriented Vision-and-Language Navigation (VLN). 
Existing methods often make decisions based on historical information, overlooking the future implications and long-term outcomes of the actions. In contrast, we aim to develop a foresighted agent. 
Specifically, we draw upon Q-learning to train a Q-model using large-scale unlabeled trajectory data, in order to learn the general knowledge regarding the layout and object relations within indoor scenes. This model can generate a Q-feature, analogous to the Q-value in traditional Q-network, for each candidate action, which describes the potential future information that may be observed after taking the specific action. 
Subsequently, a cross-modal future encoder integrates the task-agnostic Q-feature with navigation instructions to produce a set of action scores reflecting future prospects. These scores, when combined with the original scores based on history, facilitate an A*-style searching strategy to effectively explore the regions that are more likely to lead to the destination. 
Extensive experiments conducted on widely used goal-oriented VLN datasets validate the effectiveness of the proposed method. 
Our codes are available at \href{https://github.com/woyut/NavQ\_ICCV25}{https://github.com/woyut/NavQ\_ICCV25}.
\vspace{-6mm}
\end{abstract}    
\section{Introduction}
\label{sec:intro}

\begin{figure*}[t]
\begin{center}
   \includegraphics[width=\linewidth]{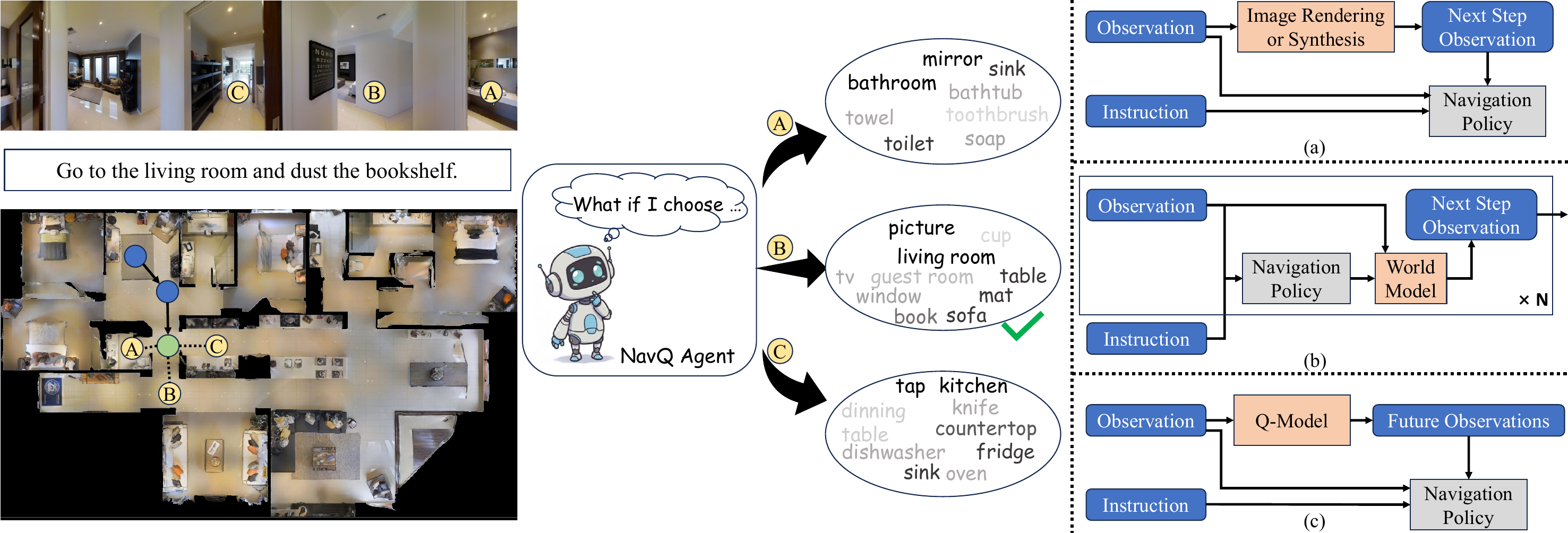}
\end{center}
\vspace{-5mm}
\caption{Left: the motivation of the proposed method. We generate cumulative Q-feature for each candidate action, which represents the future outcomes of choosing the action and enables foresighted navigation decisions. Right: a high-level comparison among the decision making processes of (a) methods based on imagining neighborhood observations, (b) methods based on a world model, and (c) our proposed method. Our Q-model is capable of forecasting the long-horizon future without time-consuming rollouts.}
\label{fig:motiv}
\vspace{-5mm}
\end{figure*}

The task of Vision-and-Language Navigation (VLN) requires an agent to reach the target location in a photo-realistic environment following language instructions. As a crucial step towards embodied intelligence, this topic has recently attracted significant attention, and many related benchmarks has been published~\cite{VLN,R4R,VLNCE,VDN,REVERIE,SOON}. In particular, REVERIE~\cite{REVERIE} concentrates on goal-oriented VLN, in which the instruction contains only the description of the target object, instead of step-by-step guidance. This setup is well-suited for the development of practical home assistants, where humans only need to provide intent-level cues rather than detailed navigation steps.

From a high-level perspective, goal-oriented VLN can be viewed as a searching problem in the scene. Despite significant progress, existing methods~\cite{HAMT,DUET,GOAT,volumetric,BEVSG} often rely solely on the information from the visited areas to make a single-step decision, without considering the potential consequences of the action.
As suggested by the A* algorithm~\cite{Astar}, integrating a heuristic metric that evaluates the future outcome when selecting frontiers to explore may greatly improve the efficiency of searching. Thus, we hope to devise a foresighted navigation agent that explicitly reason about the future prospects, in addition to the observation along the partial trajectory. A motivating example is illustrated in Figure~\ref{fig:motiv}.

Currently, several research efforts have already incorporated future information into the decision-making process, and most of them focus on predicting single-step outcomes~\cite{SIG,lookahead,FTM,ImagineNav}. By leveraging the overlapping fields of view between adjacent viewpoints, these methods can predict the scenario of the area reached after an action is taken. However, they focus on imagining the visual observations of neighboring nodes and only consider local hints, failing to capture long-range, semantic-level future information. 
On the other hand, ~\cite{LookLeap,Pathdreamer,Dreamwalker} learn a world model to predict future information in a more principled way. Though these methods can anticipate future states for any number of steps ahead, they require multi-rounds, multi-steps expansions through the world model for each decision. This rollout process is highly time-consuming and prone to distortions and overfitting, particularly when predictions are made in the RGB space~\cite{Pathdreamer,Dreamwalker}.

To address this dilemma between the horizon and efficiency, we propose NavQ, an agent that predicts the long-horizon future information within a single forward pass. At its core is a Q-model capable of anticipating the aggregated future outcomes in the latent space.
Traditionally, Q-learning will formulate a Q-function that evaluates the cumulative reward value of a state-action pair. Here, our Q-model instead outputs a Q-feature, which encapsulates the cumulation of future observations following the execution of an action. Free from reward computation, our Q-model can be pre-trained on abundant unsupervised trajectory data to enhance generalizability.
Following the Q-model, an additional cross-modal encoder is introduced to interact the Q-feature of each possible action with the text instruction, producing future-sensitive scores to complement the original decision making process based on history and current observation.

To sum up, our contributions are as follows:
\begin{itemize}
    \item We devise a Q-model that learns to predict long-range future semantics in the form of an aggregated Q-feature. We put forward a self-supervised learning pipeline to train this model on randomly-sampled trajectories without instruction annotation.
    \item We build a cross-modal future encoder that translates the Q-feature into goal-oriented heuristics. Integrating this module into a baseline model, we achieve an A*-inspired agent that makes a balance between historical progress and future prospects.
    \item Extensive experiments are conducted to demonstrate the effectiveness of the proposed method.
\end{itemize}

\section{Related Works}
\label{sec:related}
\subsection{Vision-and-Language Navigation}
Since its introduction in~\cite{VLN}, vision-and-language navigation~\cite{R4R,REVERIE,SOON,VLNCE,RxR,landmark-rxr} has received significant attention in recent years. Existing works address this task through various approaches, including (1) the exploration of different learning strategies such as imitation learning~\cite{LPSS}, reinforcement learning~\cite{LookLeap,RCM,SERL,CRL}, adversarial training~\cite{CMGAAL,counterfactual,adversarial}, generative modeling~\cite{generative}, curriculum learning~\cite{Curriculum}, cycle-consistent learning~\cite{CCC}, and energy-based optimization~\cite{EBP}; (2) the design of offline pre-training~\cite{TRL,generic,VLNBERT,SIA,AirBERT,NaViT,HOP,GELA}, auxiliary tasks~\cite{SelfMonitor,AuxReason,LAD,LANA,CITL}, and regularizations~\cite{EnvAgnostic,counterfactual2,GOAT} for a more stable and less biased training process; (3) the development of more informative history representations and scene representations~\cite{ghost,EGP,topo,RBERT,SSM,HAMT,volumetric,RSSE,MTVM,DSRG,BEVBert,BEVSG,GridMM};  (4) the design of action space for efficient exploration and backtracking~\cite{TacRewind,regret,active,DUET,MetaExplore,AZHP}; (5) the extraction of finer-grained visual~\cite{looking,EnvBias,ORIST,SOAT,AACL,GeoVLN} and textual features~\cite{BabyWalk,OAAM,NVE,LVER,SubInstruction,syntactic,VLN-Trans,ADAPT,DDL} or the incorporation of external knowledge from large language models (LLM)~\cite{MiC,LLMCopilot,NavGPT,MapGPT,NaVid,NavGPT2,correctable,NaviLLM,LangNav,Discuss}, vision-language models (VLM)~\cite{LAD}, and knowledge bases~\cite{CKR,KERM,ACK}; (6) the implementation of data augmentation techniques, including observation perturbation~\cite{EnvEdit,SESIS,FDA,DevRobust}, automatic trajectory annotation~\cite{SF,MMD,EnvDrop,LessMore,PES,SRI,CNIG,NIGBEV,flywheel} and creating new scenes~\cite{EnvMixup,newPath,unlabel,ScaleVLN,youtube,PanoGen}; and (7) the introduction of diverse related tasks~\cite{VNLA,VDN,HANNA,JustAsk,RMM,SCoA,RobotSlang,VxN} and practical settings~\cite{VLN-PETL,FastSlow,sim2real,IVLN,GSA}.

In particular, a line of works focusing on leveraging future information offers inspirations for our method. Existing attempts can be roughly classified into three paradigms. (1) Some of them~\cite{LookLeap,Pathdreamer,Dreamwalker} train a generative world model that outputs the next observation given current observation and an action. With this model, candidate actions can be mentally expanded for multiple steps (using beam search or MCTS), and the consistency of the resulting paths with the text instruction is used to evaluate the corresponding action.
(2) Other works employ future-related information to augment the visual features. \cite{SIG,lookahead,FTM} leverage various techniques like dVAE, volume rendering, NeRF, or diffusion to synthesize the resulting observation of an action. Upon the synthesized images, \cite{ImagineNav} further consults a VLM to reason about them. 
(3) Also, there is a series of attempts~\cite{TDSTP,ImagineGo,PEANUT,SSCNav,ASGN,PONI}, mainly in Object Navigation (ObjNav), working on completing the unobserved area or predicting a possible sub-goal in a top-down map. 
Different from the works above, our method directly predicts the Q-feature of each candidate action, thus it does not involve the time-consuming step-by-step rollout of a world model (in contrast to (1)) nor the explicit construction of a metric map (in contrast to (3)). On the other hand, we focus on the long-horizon, high-level, heuristic future semantics rather than he immediate, localized, reconstruction-based neighborhood information (in contrast to (2)).

\subsection{Q-Learning and Q* Agent}
As a classic algorithm in reinforcement learning (RL), Q-learning~\cite{Q} and its deep learning-based variants~\cite{DQN,DoubleQ,Rainbow} have achieved breakthroughs in game playing and beyond. 
Later, there has been a growing body of research exploring the integration of Q-learning with the powerful representational capabilities of Transformers~\cite{Qtrans,Q-DT,StopRegress}, leading to notable advancements in the field of embodied intelligence. 
More recently, the concept of the Q* algorithm has garnered remarkable attention, especially in the realm of LLM-based reasoning and planning. 
\cite{Q*} combines Q-learning with A* search~\cite{Astar} to improve the multi-step reasoning capability of the LLM. It proposes to learn a Q-value model on sampled reasoning trajectories, and the output of this model is added with a process-based reward to determine the best action at each step.
\cite{AgentQ} and \cite{Q*Agent} also estimate the Q-value of the agent's actions, which then serves as feedbacks and enables the self-improvement of LLM. 
In this work, we also aim to employ a combination of Q-learning and A* search. However, instead of building a general inference pipeline for LLMs, we design a grounded agent in the specific context of navigation. We borrow the idea of A* to implement a foresighted embodied agent, while leveraging Q-learning to efficiently equip the agent with knowledge on future outcomes.

An ObjNav method VLV~\cite{VLV} also involves Q-learning for navigation. It learns a value function from YouTube videos that outputs the Q-value for an image-action pair, representing the closeness to certain object classes. 
It should be noted that this method cannot be trivially adapted to our task, as the target in VLN is not specified by a closed-set object category. Instead, by advancing from Q-value to Q-feature, we manage to capture general-purpose, target-agnostic, future-centric knowledge from unlabeled paths. Thus, the model design and training process of our Q-model diverge significantly from that of VLV.

\section{Method}
\subsection{Task Setting and Base Model}
\begin{figure}[t]
\begin{center}
   \includegraphics[width=\linewidth]{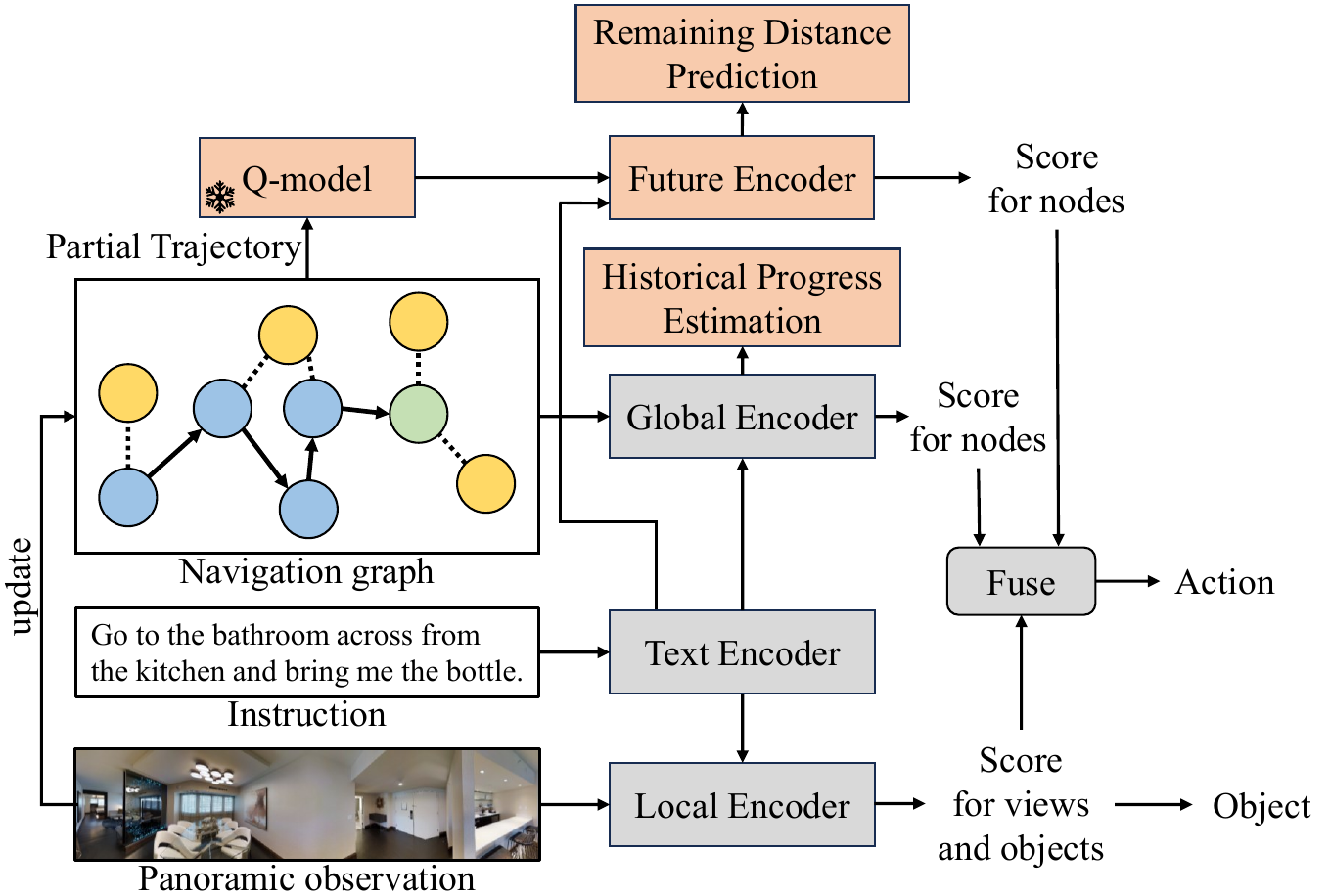}
\end{center}
\vspace{-5mm}
\caption{An overview of the proposed model. The gray modules are inherited from the baseline model~\cite{DUET}, while the orange ones are introduced by this work.}
\label{fig:method}
\vspace{-5mm}
\end{figure}

The target of goal-oriented VLN, or Remote Embodied Visual referring Expression, is to navigate to an object referred by text instruction. The reachable places in the scene are abstracted as a graph. At each time step, the agent perceives a panoramic image at its current node, and selects a neighboring node as its action. The panoramic observation is usually divided into 36 discrete views. We use DUET~\cite{DUET} as the base agent. As shown in Figure~\ref{fig:method}, it maintains a graph of the visited nodes and candidate nodes (the nodes that have been observed but not visited) during the navigation process. When determining actions, it interacts the textual feature with the coarse-grained node features on the graph and the fine-grained view features around the current position, using a global encoder (GE) and a local encoder (LE), respectively. The resulting dual-scale features are fused together to predict the action scores for all the candidate nodes on the graph. 
Formally, the major computation process of DUET can be summarized as:
\begin{align}
\label{eq:duet-graph}
G^t & = \text{Update}(\left\{r_i^t\right\}_{i=1}^{N},G^{t-1}), \\
\label{eq:duet-GE}
\left\{\hat{v}^t_i\right\}_{i=0}^{|G^t|} & = \text{GE}(G^t, w), \\
\label{eq:duet-LE}
\left\{\hat{r}^t_i\right\}_{i=1}^{N}, \left\{\hat{o}^t_i\right\}_{i=1}^{M^t} & = \text{LE}(\left\{r^t_i\right\}_{i=1}^{N}, \left\{o^t_i\right\}_{i=1}^{M^t}, w), \\
\label{eq:duet-fuse} 
p^{a,t} & = \text{Fuse}(\left\{\hat{v}^t_i\right\}_{i=0}^{|G^t|}, \left\{\hat{r}_i\right\}_{i=1}^{N}), \\ 
p^{o,t} & = \text{Pred\_Obj}(\left\{\hat{o}^t_i\right\}_{i=1}^{M^t}).
\end{align}
At timestep $t$, $\left\{r^t_i\right\}$ are the image features of the $N=36$ views at current location, $G^t$ is the maintained graph, $w$ is the feature of the text instruction, $\left\{o^t_i\right\}$ are the features of $M^t$ possible objects at current location. The output $p^{a,t}$ and $p^{o,t}$ are probability distributions over the candidate nodes and the possible objects, respectively.

\begin{figure*}[t]
\begin{center}
   \includegraphics[width=\linewidth]{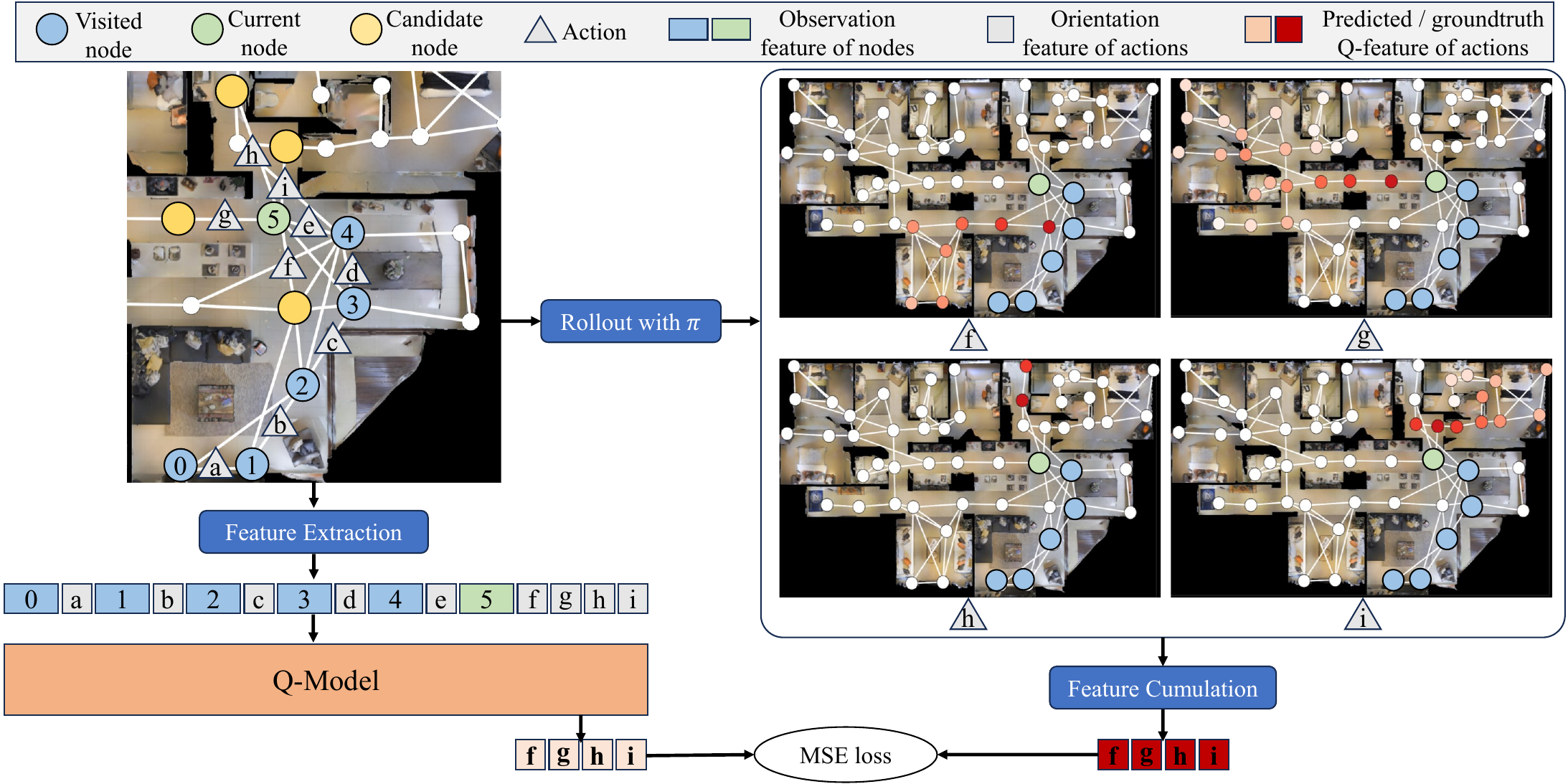}
\end{center}
\vspace{-5mm}
\caption{The design of our Q-model. Given a randomly sampled partial trajectory, the Q-model takes the observation and action features along the way as input, and predicts the Q-features for the candidate actions (f, g, h, i) at current node. The ground-truth Q-feature is the cumulated feature of all possible future nodes. We present a visualization of the cumulated nodes for each candidate action. The intensity of the red color on the node reflects the magnitude of the decay factor ($\gamma^t$) for cumulation.}
\label{fig:Q}
\vspace{-5mm}
\end{figure*}

\subsection{Overview}
Since the action scores produced by DUET are purely based on history information in the explored area, we propose to introduce an additional future-related branch into the pipeline, running in parallel with GE. As illustrated in Figure~\ref{fig:method}, the added branch comprises a Q-model and a future encoder. The former generates Q-features for each navigation candidate, aggregating potential future observations along that direction into a latent vector. The latter further utilizes the text instruction to transform the task-agnostic Q-features into scores that are helpful for the navigation problem. 
Intuitively, by integrating the anticipation of long-horizon outcomes into the decision-making process, the model is expected to select more foresighted and efficient navigation actions. 
In the following two subsections, we will detail the design and training of these two modules.

\subsection{Q-Model}
\label{sec:method-Q}

In RL, the Q-value of a state-action pair is defined as the expected cumulative reward that an agent can attain by taking the action. A high-quality value function will help the agent execute prescient decision-making and select the optimal action. To train such a value function, we first need to define an appropriate reward function. For VLN, rewards are naturally related to the destination described by natural language instructions (e.g., distance to the destination)~\cite{SERL,LookLeap,RCM}. However, the scarcity of instruction-annotated data stands as a notorious issue in this field, prompting a series of endeavors~\cite{SF,MMD,EnvDrop,LessMore,PES,SRI,CNIG,NIGBEV} trying to alleviate it. In light of this, we hope to decouple the reward computation from the training of our Q-model, by making the Q-model estimate future observations rather than future rewards. In particular, we define the Q-function as follows:
\begin{align}
\label{eq:Q}
\boldsymbol{Q}(\mathbf{T},a)=\boldsymbol{R}(\mathcal{A})+\gamma\mathbb{E}_{a'\sim\pi(a'|\mathbf{T}\cup\left\{\mathcal{A}\right\})}[\boldsymbol{Q}(\mathbf{T}\cup\left\{\mathcal{A}\right\},a')]
\end{align}
In the context of VLN, the state is a partial trajectory $\mathbf{T}$. The action $a$ is to choose a local candidate node $\mathcal{A}$, which will deterministically lead to a new state $\mathbf{T}\cup\left\{\mathcal{A}\right\}$. $\gamma$ is the decay ratio. $\boldsymbol{R}$ is a feature extractor. $\pi$ is a navigation policy. 
It is clear that the formulation of Eq (\ref{eq:Q}) is irrelevant with the navigation destination and the trajectory description, so $\boldsymbol{Q}$ can be learned on annotation-free scenes.

\subsubsection{Data Gathering and Supervision}
\label{sec:method-Q-data}
Based on the Bellman equation, classical Deep Q-Learning (DQN)~\cite{DQN-nature} updates the Q-model using the gradients of temporal difference errors.
In VLN, since there are finite nodes on the graph and revisiting is prohibited, the recursion in Eq (\ref{eq:Q}) will end at some point where the current node has no valid candidate. Actually, the probability of reaching a particular node from a state-action pair under policy $\pi$, as well as the distribution of the number of steps required to reach it, can be calculated by enumerating all feasible episodes (i.e., terminated trajectories) on the graph. To be specific,
\begin{align}
\label{eq:pi-prob}
\text{P}_{\pi}(\mathcal{N},t|\mathbf{T}, a)=\sum_{\tilde{\mathbf{T}}\in\mathbb{T}(\mathbf{T},\mathcal{A},\mathcal{N},t)}\text{P}_{\pi}(\mathbf{T}\cup\left\{\mathcal{A}\right\}\rightarrow\tilde{\mathbf{T}}).
\end{align}
Here, $\mathbb{T}(\mathbf{T},\mathcal{A},\mathcal{N},t)$ is the set of terminated trajectories that satisfy: (i) each trajectory starts with $\mathbf{T}\cup\left\{\mathcal{A}\right\}$, (ii) the portion of the trajectory after $\mathcal{A}$ contains node $\mathcal{N}$, and (iii) the number of steps from $\mathcal{A}$ to $\mathcal{N}$ is $t$. $\text{P}_{\pi}(\cdot\rightarrow\cdot)$ is the probability of expanding a partial trajectory into a complete trajectory under the policy $\pi$. With this distribution of nodes and steps, Eq (\ref{eq:Q}) can be transformed into a more straightforward equation:
\begin{align}
\label{eq:Q-sup}
\boldsymbol{Q}(\mathbf{T},a)=\sum_{\mathcal{N},t}\text{P}_{\pi}(\mathcal{N},t|\mathbf{T}, a)\gamma^t\boldsymbol{R}(\mathcal{N}).
\end{align}
This formulation provides the ground-truth Q-feature for any state-action pair, enabling us to train our Q-model without RL techniques.

The rollout policy $\pi$ determines the characteristic of the learned Q-function. A naive idea is to set it as a random policy that uniformly chooses a candidate as action. However, such a design can lead to a lack of discrimination between different candidates. In the graphs of navigation scenes, there are often numerous loops, which means that for an unexplored node $\mathcal{N}$, multiple candidate actions from the current node may potentially lead to it (i.e., $\mathbb{T}(\mathbf{T},\mathcal{A},\mathcal{N},t)$ is non-empty for multiple candidate nodes $\mathcal{A}$). As a result, $\boldsymbol{R}(\mathcal{N})$ will be accumulated into the Q-features of multiple candidate actions, making them less informative. Put in another way, random exploration is highly inconsistent with the actual strategy adopted by a normal VLN model.
To handle this problem, we note that goal-oriented VLN aims at finding the most efficient way to reach a target object. The best trajectory for any instruction is always the shortest path between two nodes. Based on this, we aim to incorporate a preference for the optimality of the future paths into the policy. We achieve this by introducing a fourth condition into the definition of the path set $\mathbb{T}(\mathbf{T},\mathcal{A},\mathcal{N},t)$ in Equation (\ref{eq:pi-prob}): in each trajectory $\tilde{\mathbf{T}}$, the segment from $\mathbf{T}[-1]$ to $\tilde{\mathbf{T}}[-1]$ must be a shortest path. With this additional requirement, it can be proven that for any partial trajectory $\mathbf{T}$ and any node $\mathcal{N}$, there exists at most one pair of $(a,t)$ that satisfies $\text{P}_{\pi}(\mathcal{N},t|\mathbf{T}, a)>0$. This implies that the feature of each possible future node is accumulated into the Q-feature of a single action through a unique path.
Figure~\ref{fig:Q} illustrates the sets of future nodes accumulated to different action candidates, along with the corresponding rollout steps $t$ for each node.
This policy design enables the learned Q-features to comprehensively aggregate diverse future observations while reflecting differences in navigation efficiency across actions, achieving a balance between coverage and optimality.

To sum up, the data generation pipeline for training our Q-model is: (i) sample a trajectory $\mathbf{T}$ of arbitrary length in the scene; (ii) sample an action $a$ at the last node of the trajectory; (iii) use Eq (\ref{eq:Q-sup}) to compute the ground-truth Q-feature as supervision.

\begin{table*}[]
\caption{The results on REVERIE. The best and second-best results are marked as \textbf{bold} and \underline{underline}, respectively.}
\vspace{-5mm}
\label{tab:main}
\begin{center}

\scalebox{0.85}{
\begin{tabular}{l|ccccc|ccccc}
               & \multicolumn{5}{c|}{Val Unseen} & \multicolumn{5}{c}{Test Unseen} \\ 
               & OSR$\uparrow$ & SR$\uparrow$ & SPL$\uparrow$ & RGS$\uparrow$ & RGSPL$\uparrow$ & OSR$\uparrow$ & SR$\uparrow$ & SPL$\uparrow$ & RGS$\uparrow$ & RGSPL$\uparrow$ \\ \hline

HAMT~\cite{HAMT} \tiny{[NeurIPS21]}  &36.84 &32.95 &30.20 &18.92 &17.28  &33.41 &30.40 &26.67 &14.88 &13.08\\ 
HOP~\cite{HOP} \tiny{[CVPR22]} &36.24 &31.78 &26.11 &18.85 &15.73  &33.06 &30.17 &24.34 &17.69 &14.34\\ 
LANA~\cite{LANA} \tiny{[CVPR23]}  &52.97 &48.31 &33.86 &32.86 &22.77  &57.20 &51.72 &36.45 &32.95 &22.85\\
AZHP~\cite{AZHP} \tiny{[CVPR23]} & 53.65 & 48.31 & 36.63 & 34.00 & 25.79  & 55.31 & 51.57 & 35.85 & 32.25 & 22.44 \\
KERM~\cite{KERM} \tiny{[CVPR23]} & 55.21 & 50.44 & 35.38 & 34.51 & 24.45  & 57.58 & 52.43 & 39.21 & 32.39 & 23.64 \\
BEV-Bert~\cite{BEVBert} \tiny{[ICCV23]}&- &51.78 &36.37 &34.71 &24.44 &- &52.81 &36.41 &32.06 &22.09\\
BSG~\cite{BEVSG} \tiny{[ICCV23]}& 58.05 & 52.12 & 35.59 & 35.36 & 24.24 & \textbf{62.83} &56.45 &38.70 & 33.15 &22.34\\
GridMM~\cite{GridMM} \tiny{[ICCV23]}& 58.48  &51.37 &36.47 &34.57 &24.56  &59.55 &53.13 &36.60 &\underline{34.87} &23.45\\
FDA~\cite{FDA} \tiny{[NeurIPS23]} & 51.41 & 47.57 & 35.90 & 32.06 & 24.31 & 53.54 & 49.62 & 36.45 & 30.34 & 22.08\\
GOAT~\cite{GOAT} \tiny{[CVPR24]} & - & \underline{53.37} & 36.70 & \textbf{38.43} & \underline{26.09} & - & \textbf{57.72} & \textbf{40.53} & \textbf{38.32} & \textbf{26.70} \\
VER~\cite{volumetric}  \tiny{[CVPR24]} & \textbf{61.09} & \textbf{55.98} & \textbf{39.66} & 33.71 & 23.70 & \underline{62.22} & \underline{56.82} & 38.76 & 33.88 & 23.19 \\ 
ENP~\cite{EBP} \tiny{[NeurIPS24]} &54.70 &48.90 &33.78 &34.74 &23.39  &59.38 &53.19 &36.26 &33.10 &22.14 \\ 
 \hline
baseline~\cite{DUET} \tiny{[CVPR22]} & 51.07 & 46.98 & 33.73 & 32.15 & 23.03 & 56.91 & 52.51 & 36.06 & 31.88 & 22.06 \\
NavQ & \underline{60.47} & 53.22 & \underline{38.89} & \underline{36.84} & \textbf{27.12} & 60.39 & 53.29 & \underline{39.50} & 34.82 & \underline{25.16} \\
& (+9.40) & (+6.24) & (+5.16) & (+4.69) & (+4.09) & (+3.48) & (+0.78) & (+3.44) & (+2.94) & (+3.10) \\
\hline
 \textit{Methods with additional scenes} \\ \hline
AutoVLN~\cite{unlabel} \tiny{[ECCV22]} & \textbf{62.14} & \underline{55.89} & \underline{40.85} & \underline{36.58} & \underline{26.76} & \textbf{62.30} & \underline{55.17} & 38.88 & 32.23 & 22.68 \\
Lily~\cite{youtube} \tiny{[ICCV23]} & 53.71 & 48.11 & 34.43 & 32.15 & 23.43 & 60.51 & 54.32 & 37.34 & 32.02 & 21.94 \\
ScaleVLN~\cite{ScaleVLN} \tiny{[ICCV23]} & - & \textbf{56.97} & \textbf{41.84} & 35.76 & 26.05  & - & \textbf{56.13} & \underline{39.52} & \underline{32.53} & \underline{22.78} \\
PanoGen~\cite{PanoGen} \tiny{[NeurIPS23]} & - & 51.18 & 34.99 & 33.26 & 22.99 & - & - & - & - & - \\\hline

NavQ (w.o. speaker annotation)& \underline{62.00} & 54.10 & 39.22 & \textbf{37.57} & \textbf{27.29} & \underline{61.25} & 54.91 & \textbf{40.08} & \textbf{35.87} & \textbf{25.14}\\
\end{tabular}
}
\vspace{-8mm}
\end{center}
\end{table*}

\subsubsection{Model and Training}
\label{sec:method-Q-train}
The Q-model is designed as a Transformer. As shown in Figure~\ref{fig:Q}, the input consists of interleaved node features and action features of the partial trajectory, followed by the features of candidate actions at current location. Multiple candidates can be processed in a single forward pass as they share the trajectory prefix. We use the set of view features, $\left\{r_i^t\right\}$, as the descriptor of each node, while the actions are encoded by $\text{sin}$ and $\text{cos}$ values of the orientations. 
The outputs corresponding to the candidate tokens are used as predicted Q-features. MSE loss between the predictions and the ground-truth is employed to train the network.

The key consideration in pre-training the Q-model is to achieve generalizability. The model is expected to learn the common patterns regarding room layouts and object placements, rather than simply memorizing the details of the training scenes. Using large-scale random trajectories for training can solve this problem to some extent. Yet, due to the limited number of training scenes, the model is still at risk of overfitting. We further alleviate this issue through the following designs.

\textbf{(1) Text-based Prediction}. The visual features of RGB views inevitably carry some stylistic and texture information, which is usually unrelated to the navigation task. The Q-model trained on these features may establish some spurious correlations, making it difficult to generalize to new scenes. We propose to learn the Q-model in the latent text space, \textit{i.e.,} the feature extractor $\boldsymbol{R}$ is designed to be the feature of the natural language description of a node. These descriptions can be obtained by pre-processing the scenes with an off-the-shelf image captioning model~\cite{BLIP,LangNav}. By predicting the abstracted text-based features of future observations, the Q-model can better focus on high-level semantic relationships, thereby providing more reliable guidance for the navigation task.

\textbf{(2) Warm-up Pre-training}. Self-supervised pre-training is proven beneficial in many vision and language tasks. Before performing regression on the Q-features, we first carry out an MAE pre-training~\cite{MAE}. The input format is the same as described above, with some randomly selected tokens set to zero. An additional MLP is appended after the Transformer to reconstruct the masked tokens. This training process provides a good initialization for the Q-training and guides the model to fully analyze the information in the trajectory history.

\subsection{Future Encoder}
\label{sec:method-FRM}
With the Q-model at hand, we can generate Q-features for the candidate actions at each navigation step, representing the scenarios the agent may encounter after it takes the action. 
The future encoder (FE) is responsible for transforming the task-agnostic feature into goal-oriented information. Formally,
\begin{align}
\left\{\hat{q}^t_i\right\}_{i=0}^{|\tilde{G}^t|} & = \text{FE}(\tilde{G}^t, w).
\end{align}
$\tilde{G}^t$ is a graph with the same topology as $G^t$ (Eq (\ref{eq:duet-graph})), while it is updated with the Q-features of the candidate nodes instead of the view features. FE is designed as a Graph Transformer that shares the same architecture as GE. The output $\left\{\hat{q}^t_i\right\}_{i=0}^{|\tilde{G}^t|}$ is integrated into the fusion process described in Eq (\ref{eq:duet-fuse}).

Ideally, the GE branch is tasked with analyzing historical information, while the FE branch handles future information. To ensure this decomposition and improve the performance of each branch, we introduce some additional supervisory signals.
In previous works, progress monitor~\cite{SelfMonitor} is a widely-used auxiliary task, which requires the model to predict at each timestep the progress it has made towards the destination. Here we adopt this idea and designs two progress-related subtasks. For each node, on one hand, we send GE's node feature $\hat{v}^t_i$ to a lightweighted MLP to predict the traversed distance up to now. On the other hand, we send FE's output $\hat{q}^t_i$ to another MLP to predict the remaining distance to go. The ground-truth for them are designed as:$s_1(\mathcal{A}) = (\text{dist}(\mathcal{S},\mathcal{C})+\text{dist}(\mathcal{C},\mathcal{A}))/D_1, s_2(\mathcal{A}) = \text{dist}(\mathcal{A},\mathcal{G})/D_2$,
where $\mathcal{S}$, $\mathcal{C}$, and $\mathcal{G}$ are the starting, current, and goal nodes, $\text{dist}(\cdot)$ is the shortest distance between two nodes, $D_1$ and $D_2$ are normalizing constants.
The combination of these two sub-tasks also reflects the idea of integrating the cost function with a goal-directed heuristic function in the A* algorithm~\cite{Astar}, allowing the future information embedded in the Q-feature to be effectively utilized by the navigation agent.

\subsection{Training Scheme}
\label{sec:method-train}
The training process of NavQ is divided into three stages.

\textbf{Stage 1: Q-model pre-training}. As detailed in Section~\ref{sec:method-Q}, we first pre-train the Q-model on randomly sampled trajectories in the training scenes. 
The Q-model will be kept frozen and used as a feature extractor in the following stages. 

\textbf{Stage 2: Agent pre-training}. Pre-training on offline instruction-trajectory pairs is proven effective by many recent works~\cite{AirBERT,RBERT,DUET,generic,HOP}. We adopt the four pre-training tasks implemented by DUET. 
Besides, to give direct guidance to FE and GE, the two progress-related tasks mentioned in Section~\ref{sec:method-FRM} are also included. 
Details of these tasks can be found in the supplementary material.

\textbf{Stage 3: Agent finetuning}. We still follow DUET to finetune the agent on online data using DAgger~\cite{dagger} wth a pseudo expert policy.

\section{Experiments}
\subsection{Datasets and Metrics}
Experiments are performed on two popular VLN benchmarks, REVERIE~\cite{REVERIE} and SOON~\cite{SOON}. Both are goal-oriented VLN datasets based on the MP3D simulator~\cite{VLN}, requiring the agent to navigate to a target object instance. 
REVERIE includes a set of high-level instructions that guide the agent toward the target object located 4 to 7 steps away.
SOON is a more challenging dataset with longer target descriptions and an average trajectory length of 9.5. 
We evaluate the model on the official validation set and test set, both consisting of previously unseen scenes during training.
The metrics include success rate (SR), oracle SR (OSR), SR penalized by path length (SPL), remote grounding success (RGS), RGS penalized by path length (RGSPL). Detailed descriptions of the datasets and metrics can be found in the supplementary material. 

\subsection{Implementation Details}
The Q-model is implemented as a 4-layer Transformer, and the FE is a 4-layer Graph Transformer. The remaining parts of the model follow the same architecture as DUET~\cite{DUET}. 
The batch size, learning rate, and iterations for the three training stages are set to 128/32/4, 1e-5/5e-5/1e-5, 30k/100k/20k, respectively. CLIP-ViT/B is used as the visual and textual feature extractors for its cross-modal performance. The training can be conducted on a single NVIDIA RTX 3090 GPU. More details are presented in the supplementary material.

\subsection{Main Results}

\begin{table}[]
\caption{The results on SOON. The best and second-best results are marked as \textbf{bold} and \underline{underline}, respectively.}
\vspace{-5mm}
\label{tab:main-soon}
\begin{center}

\scalebox{0.85}{
\begin{tabular}{c|l|cccc}
               & & OSR & SR & SPL & RGSPL \\ \hline
Val & GBE~\cite{NIGBEV} &28.54 &19.52 &13.34 &1.16\\
Unseen & GridMM~\cite{GridMM} &53.39 &37.46 &24.81 &3.91 \\
& KERM~\cite{KERM} & 51.62 &38.05 &23.16& 4.04 \\
& AZHP~\cite{AZHP} & \underline{56.19} &\textbf{40.71} &26.58& \underline{5.53} \\
& GOAT~\cite{GOAT}  & 54.69 & \underline{40.35} & \textbf{28.05} & \textbf{6.10} \\ \cline{2-6}
& baseline~\cite{DUET} & 50.91 & 36.28 & 22.58 & 3.75 \\ 
& NavQ (Ours) & \textbf{58.79} & 39.09 & \underline{26.65} & 5.51 \\
&  & (+7.88) & (+2.81) & (+4.07) & (+1.76)

\\ \hline
Test & GBE \cite{NIGBEV}  &21.45 &12.90 &9.23 &0.45\\
Unseen & GridMM \cite{GridMM}&48.02 &36.27 &21.25 &4.15\\
& GOAT~\cite{GOAT}  & \textbf{50.63} & \textbf{40.50} & \textbf{25.18} & \textbf{6.10} \\ \cline{2-6}
& baseline~\cite{DUET} & 43.00 & 33.44 & 21.42 & 4.17\\
& NavQ (Ours) &  \underline{48.92} & \underline{38.59} & \underline{24.50} & \underline{4.48}\\
&  & (+5.92) & (+5.15) & (+3.08) & (+0.31)\\
\end{tabular}
}
\vspace{-10mm}
\end{center}
\end{table}

Table~\ref{tab:main} shows the performance comparison on REVERIE. Our NavQ agent consistently outperforms the DUET~\cite{DUET} baseline across all evaluation metrics, showing the effectiveness of incorporating the future branch. Compared to state-of-the-art models based on techniques such as causal learning~\cite{GOAT} and volumetric representation~\cite{volumetric}, our model also demonstrates competitive performance, \textit{e.g.}, +3.4\%/+2.0\% RGSPL than VER on the validation/test set, +2.2\% SPL than GOAT on the validation set. 
One advantage of our method is that the Q-model could benefit from training on large-scale unlabeled scenes. To prove this, we borrow scenes from the HM3D~\cite{HM3D} and Gibson~\cite{Gibson} simulator following~\cite{ScaleVLN}, and obtain a total of 1,351 scenes for Q-training. Note that we do not employ any speaker model~\cite{SF} to label the trajectories in the additional scenes, and these scenes are only used in training stage 1. As illustrated in the lower part of Table~\ref{tab:main}, using additional scenes further boosts NavQ's navigation capability, reaching a performance comparable or higher than the methods that utilize additional annotated trajectories~\cite{unlabel,youtube,ScaleVLN,PanoGen}.

Similarly, as in Table~\ref{tab:main-soon}, our model also performs better than the baseline for all metrics on SOON. 
Note that the pre-trained Q-model is shared across REVERIE and SOON. Thus, the results highlight the task-agnostic nature of the learned Q-model, and demonstrate the generalizability of our approach.

\begin{figure}[t]
\begin{center}
   \includegraphics[width=0.8\linewidth]{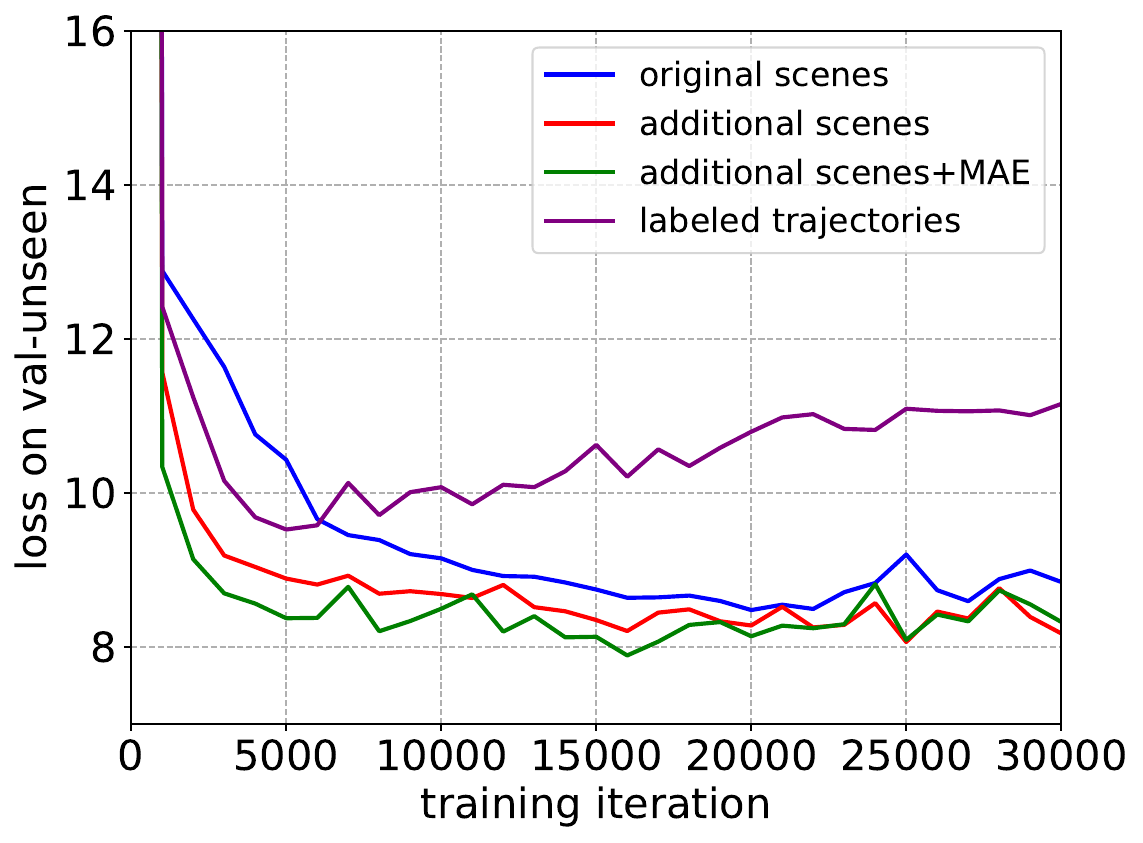}
\end{center}
\vspace{-5mm}
\caption{A comparison among different training techniques for the Q-model. We plot the MSE loss on the val-unseen scenes during the training process.}
\label{fig:Qloss}
\end{figure}

\begin{figure*}[t]
\begin{center}
   \includegraphics[width=\linewidth]{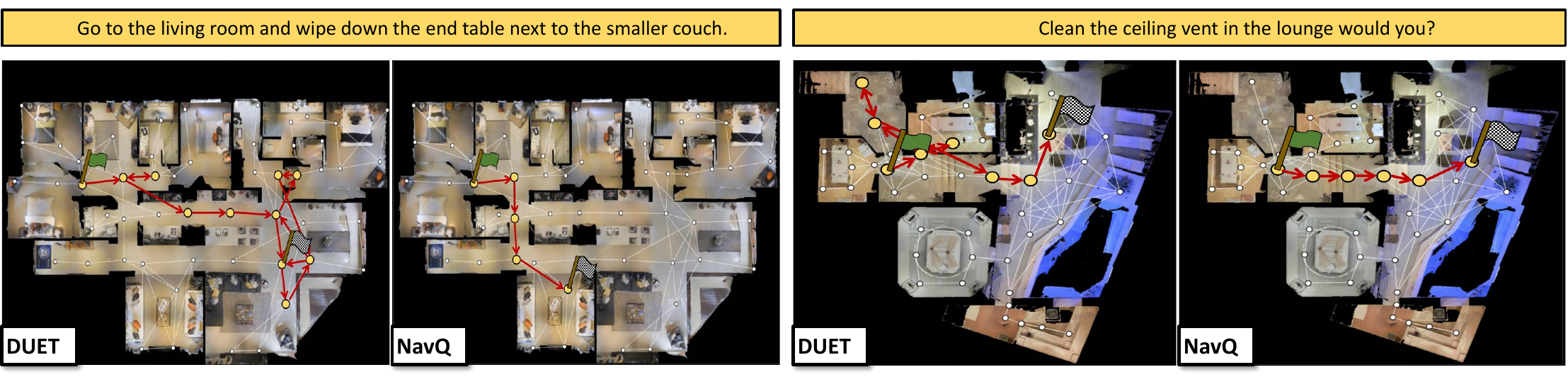}
\end{center}
\vspace{-5mm}
\caption{A qualitative comparison of our method and the baseline agent. In both examples, the NavQ agent performs the instruction correctly while the baseline agent fails.}
\label{fig:qual}
\vspace{-5mm}
\end{figure*}
\subsection{Ablation Studies}

\begin{table}[]
\caption{Ablation study on the future branch. The results are obtained on REVERIE's unseen validation set.}
\vspace{-5mm}
\label{tab:abl-future}
\begin{center}
\scalebox{0.85}{
\begin{tabular}{c|cc|ccccc}
               & QM & FE & OSR & SR & SPL & RGS & RGSPL \\ \hline
(1) & \XSolidBrush & \XSolidBrush & 54.42 & 48.14 & 33.38 & 30.19 & 21.05\\
(2) & \XSolidBrush & \Checkmark & 54.84 & 48.20 & 33.92 & 32.52 & 23.14 \\
(3) & \Checkmark & \XSolidBrush & 53.25 & 48.48 & 32.22 & 33.03 & 21.86\\
(4) & \Checkmark & w.o. loss & 55.98 & 51.55 & 35.79 & 34.51 & 23.81\\
(5) & \Checkmark & w. loss & \textbf{60.47} & \textbf{53.22} & \textbf{38.89} & \textbf{36.84} & \textbf{27.12} \\ \hline
(6) & GT & \XSolidBrush & 60.18 & 54.36 & 41.71 & 37.03 & 28.59\\
(7) & GT & \Checkmark & \textbf{65.38} & \textbf{59.27} & \textbf{47.04} & \textbf{39.68} & \textbf{31.62}\\
\end{tabular}
}
\vspace{-10mm}
\end{center}
\end{table}

We conduct an ablational experiment on the role of the Q-model (QM) and the future encoder (FE) in the future branch. The compared architectures include:
(1) A variant without the future branch, \textit{i.e.}, a reproduced version of the baseline model.
(2) A variant that utilizes FE but not QM, where FE receives the same input as GE.
(3) A variant that utilizes QM but not FE, where the output of QM is concatenated with the view features $\left\{r_i^t\right\}_{i=1}^{N}$ and fed into GE (Eq(\ref{eq:duet-graph}-\ref{eq:duet-GE})).
(4) A model that utilizes both QM and FE, but without supervision from the progress-related subtasks during the second training stage.
(5) The full NavQ model.

The results are shown in Table~\ref{tab:abl-future}.
We first notice that our reproduced baseline has higher OSR/SR but lower RGS/RGSPL than the reported performance of DUET, which may be attributed to the use of different visual backbones. (We use CLIP-ViT/B to enhance the cross-modal capability of the Q-model, while DUET employs an ViT-B/16 pre-trained on ImageNet which is the same as the object feature extractor.) 
Upon that, merely introducing FE provides only limited improvement, suggesting that leveraging historical information alone may not be sufficient.
On the other hand, the improvement achieved by solely using the Q model is minor too, indicating the significance of employing FE to extract task-relevant information from the rich future context.
The progress-related losses also contribute to the overall performance, validating the benefits of applying direct supervisions to decouple the history branch and the future branch.

In addition, we experiment with replacing the outputs of QM with the ground-truth (GT) Q-features, which serves as an upper bound for our method. 
The GT Q-features are sent to FE ((7) in Table~\ref{tab:abl-future}) or concatenated with the view features and sent to GE ((6) in Table~\ref{tab:abl-future}).
It can be observed that using the GT Q-features significantly enhances performance, especially on the metrics related to navigation efficiency (\textit{e.g.}, +14\% SPL and +11\% RGSPL over the baseline).
These results validate the design of the Q-feature (Eq (\ref{eq:Q-sup})) and the choice of the rollout policy $\pi$ that incorporates a preference for shortest paths. 
Also, the superiority of using FE remains valid when high-quality Q-features are available.

To take a closer look at the training process of the Q-model, we plot the curve of validation loss on the scenes in the val-unseen set. As shown in Figure~\ref{fig:Qloss}, training the Q-model with randomly sampled paths achieves better generalization than training solely with annotated paths, due to the vast difference in the number of training samples. This observation is a key factor motivating us to design a Q-learning paradigm without instruction annotations. Meanwhile, introducing additional training scenes and adding the MAE pre-training for Q-model also show positive influence on the quality of Q-features, which in turn leads to better navigation performance as in Table~\ref{tab:main}.

Besides, we also conduct an analysis on the decay ratio $\gamma$, which is a key hyper-parameter in the design of Q-model. When $\gamma=0$, the Q-model is reduced to only predicting the observation of the immediate next step, like a world model or a novel view synthesis model. As $\gamma$ grows larger, the ground-truth Q-features will encompass richer future information. At the same time, the training of the Q-model will become more challenging, and the discrepancy between the predicted Q-feature and the ground-truth will increase. We choose $\gamma=0.5$ as a default setting, which makes a balance between the feature quality and the training difficulty. As shown in Table~\ref{tab:abl-decay}, it achieves higher overall navigation performance than using other values for $\gamma$. In particular, it clearly outperforms the $\gamma=0$ variant which only reconstructs the feature of neighboring nodes, demonstrating the essential role of long-term future information.

\begin{table}[]
\caption{Analysis on the effect of the decay ratio for Q-features. The results are obtained on REVERIE's unseen validation set.}
\vspace{-5mm}
\label{tab:abl-decay}
\begin{center}
\scalebox{0.94}{
\begin{tabular}{c|ccccc}
               $\gamma$ & OSR & SR & SPL & RGS & RGSPL \\ \hline
0 & 56.66 & 51.12 &  37.42 & 34.42 & 25.16\\
0.3 & 59.73 & 51.95 & \textbf{38.90} & 35.13 & 26.61\\
0.5 & \textbf{60.47} & \textbf{53.22} & 38.89 & \textbf{36.84} & \textbf{27.12}\\
0.7 & 57.06 & 50.89 & 36.15 & 33.48 & 23.85\\ 
\end{tabular}
}
\vspace{-10mm}
\end{center}
\end{table}

\subsection{Qualitative Results}
In Figure~\ref{fig:qual}, we visualize the trajectories predicted by the model on top-down floor maps. Thanks to the informative Q-features, our method can find the correct direction to explore when the items mentioned in the instruction are not yet observed. Therefore, compared to the baseline, NavQ demonstrates a higher likelihood of reaching the correct destination and exhibits greater navigation efficiency. 
\section{Conclusion}
In this work, we propose a foresighted agent for goal-oriented VLN that efficiently integrates future-relevant information into a baseline model. 
A novel Q-model is developed to represent the future outcomes of a given action in the form of aggregated features. Based on scenes without instruction annotation, we design a self-supervised training paradigm using random trajectories and put forward a series of techniques for collecting training data and enhancing model generalization.
Furthermore, we propose a future encoder that leverages instructions to transform the Q-features into assessments of candidate actions' anticipated future prospects, complementing the decision-making process that relies solely on historical information. 
In future work, we plan to further optimize the design of the Q-model, and explore extending the proposed approach to continuous environments.

\section*{Acknowledgments}

This work is supported by an internal grant of Peking University (2024JK28).

{
    \small
    \bibliographystyle{ieeenat_fullname}
    \bibliography{main}
}
\clearpage
\setcounter{page}{1}
\maketitlesupplementary

\section{Details on the Model and Training}
\subsection{Preliminaries on DUET}
\label{sec:supp-duet}
The baseline model for our NavQ agent, DUET~\cite{DUET}, proposes a dual-scale action prediction strategy on a topological graph for the VLN task. Due to its generality, DUET's architecture has been adopted by many subsequent studies~\cite{unlabel,ScaleVLN,KERM,ACK,NavGPT2,EBP,AACL}. At each navigation timestep, it involves the following computation procedures:

(1) Input processing. The agent perceives the surrounding environment at its current location through a panoramic RGB observation. The panoramic image is discretized into $N=36$ views (3 elevation angles times 12 heading angles) and processed by a frozen visual encoder. The agent also gets the feature for $M^t$ visible objects (pre-defined or detected by an off-the-shelf detector, see Section~\ref{sec:supp_datasets} for details). The feature vectors for views and objects are concatenated and sent to a learnable panorama encoding module, which is implemented as a 2-layer Transformer. The results are $\left\{r_i^t\right\}_{i=1}^{N}$ and $\left\{o^t_i\right\}_{i=1}^{M^t}$ mentioned in Eq (\ref{eq:duet-graph}) and (\ref{eq:duet-LE}). On the other hand, the word embeddings of the instruction text is sent to another 9-layer Transformer to obtain the textual feature $w$.

(2) Graph update. The agent builds a topological graph $G^t$ on the fly. The graph starts as a single node representing the starting position of the episode. VLN's task setting~\cite{VLN} assumes that the agent has access to the locations of navigable viewpoints around its current place. Thus, it can continuously expand its graph by incorporating neighboring nodes. There are two types of nodes on the graph: visited nodes and observed but unvisited nodes. For each visited node $\mathcal{N}$, the agent stores the mean of its view features ($\frac{1}{N}\sum_{i=1}^Nr_i^t$, $t$ is the step that it visits $\mathcal{N}$) as its feature. For each unvisited node $\mathcal{N}$, the agent maintains a list. If $\mathcal{N}$ is a neighboring node of the agent's location at step $t$, it identifies one of the $N$ views that is closest to the direction of $\mathcal{N}$ and inserts its feature $r_i^t$ into the list. (Note that an unvisited node can be observed by multiple visited nodes.) The agent takes the mean of this list as the feature for the corresponding node.

(3) Global action prediction. DUET features a dual-scale planning process. The coarse-scale branch outputs a probability across all the unvisited nodes on $G^t$ (\textit{i.e.}, the global candidates). It is based on a 4-layer Graph Transformer named as global encoder (GE). Each layer performs sequentially a cross-modal attention that interacts the textual feature $w$ with the node features $\left\{v^t_i\right\}_{i=0}^{|G^t|}$, and a graph-aware self-attention that takes into account the structure of the graph to further process the node features. $v^t_0$ is a zero vector representing a pseudo ``stop" node. The outputs of GE are the updated node features $\left\{\hat{v}^t_i\right\}_{i=0}^{|G^t|}$, as in Eq (\ref{eq:duet-GE}). They are transformed into logits $\left\{s^{c,t}_i\right\}_{i=0}^{|G^t|}$ by an MLP.

(4) Local action and object prediction. The fine-scale branch of DUET outputs a probability across the unvisited neighbors of the current node (\textit{i.e.}, the local candidates). It is based on a 4-layer Transformer named as local encoder (LE). Each layer performs sequentially a cross-modal attention that interacts the textual feature $w$ with the concatenation of view features $\left\{r^t_i\right\}_{i=0}^{N}$ and object features $\left\{o^t_i\right\}_{i=1}^{M^t}$, and a self-attention that further processes the concatenation. $r^t_0$ is a zero vector representing the ``stop" action. The outputs of LE are the updated view features $\left\{\hat{r}^t_i\right\}_{i=0}^{N}$ and object features $\left\{\hat{o}^t_i\right\}_{i=1}^{M^t}$, as in Eq (\ref{eq:duet-LE}). They are transformed into action logits $\left\{s^{f,t}_i\right\}_{i=0}^{N}$ and object logits $\left\{s^{o,t}_i\right\}_{i=1}^{M^t}$ by two separate MLPs.

(5) Dynamic Fusion. DUET dynamically fuses the global prediction and local prediction to get the final action logits $\left\{s^{t}_i\right\}_{i=0}^{|G^t|}$. The fusing weight is obtained by sending the concatenation of $v^t_0$ and $r^t_0$ to an MLP and a Sigmoid funtion. 

(6) Action execution. The agent selects a candidate node based on the fused probability. It then finds the shortest path from its current location to this node on $G^t$, and traverses along it. When the agent decides to ``stop", it selects an object as its prediction according to the local object probability.

\subsection{Details of the Q-Model}
\label{sec:supp-Q}
In this subsection we describe the training of the Q-model in more detail. To get each training sample, we randomly select a training scene and a starting node. Then, a partial trajectory is obtained by uniformly choosing an unvisited local candidate for a random number of steps. Based on this trajectory, the model input is formed as follows.
\begin{itemize}
    \item For each node in the trajectory, we encode its 36 view images into visual features, and pool them into 12 vectors corresponding to the 12 heading directions. We take the natural language description of each view provided by LangNav~\cite{LangNav} (extracted using BLIP~\cite{BLIP}), encode them into textual features, and pool them into 12 vectors as well. The visual features and textual features are processed by linear projections and added together, forming the full node feature of shape $12\times D$.
    \item For each action in the trajectory, we encode its orientation using $\text{sin}$ and $\text{cos}$ functions. The resulting vector is linearly projected to $D$ channels. For each local candidate actions at the current node (\textit{i.e.}, the last node of the partial trajectory), we encode it in the same way to a $D$-dimensional vector.
    \item We arrange the node features and action features alternately following the order in the trajectory, and append the candidate features at the end. As illustrated in Figure~\ref{fig:Q}, the input to the Q-model is a sequence of $13|\mathbf{T}|-1+C$ tokens, where $|\mathbf{T}|$ is the length (number of nodes) of the partial trajectory, while $C$ is the number of local candidates. Each token is a $D$-dimensional vector.
\end{itemize}

 The Q-model is implemented as a 4-layer Transformer. Apart from the traditional positional encoding that captures the order of tokens, we introduce an additional positional encoding to represent the token order within a node. Specifically, this encoding consists of 13 learnable tokens, which are added to the 13 input tokens corresponding to each node-action pair. Notably,  the last positional token is added to each candidate token.

We adopt the method described in Section~\ref{sec:method-Q-data} to form the ground-truth Q-feature. Before delving into the implementation details, we first provide a more precise formulation of the rollout policy $\pi$ used in our method. Given a partial trajectory $\mathbf{T}$ and a candidate action $a$ (leading to node $\mathcal{A}$), $\mathbf{T}\cup\left\{\mathcal{A}\right\}$ is expanded to a full trajectory $\tilde{\mathbf{T}}$ under $\pi$. At each step, the agent randomly selects a feasible local candidate according to a uniform distribution, where feasibility means that this candidate node $\mathcal{N}$ ensures that the one-step-longer rollout path is the shortest path from $\mathbf{T}[-1]$ to $\mathcal{N}$. The agent terminates when there is no feasible candidate to choose. 
This formulation is consistent with the definition of the set of possible rollout trajectories, $\mathbb{T}$.
In Section~\ref{sec:method-Q-data}, we put forward a claim that for a given pair of partial trajectory $\mathbf{T}$ and node $\mathcal{N}$, there is at most one pair of $(a,t)$ that makes $\text{P}_{\pi}(\mathcal{N},t|\mathbf{T}, a)>0$ under the policy $\pi$. This can be easily proved by contradiction. Suppose $\text{P}_{\pi}(\mathcal{N},t_1|\mathbf{T}, a_1)>0$ and $\text{P}_{\pi}(\mathcal{N},t_2|\mathbf{T}, a_2)>0$. If $t_1\neq t_2$, then there are two paths of different length going from $\mathbf{T}[-1]$ to $\mathcal{N}$. They cannot simultaneously be the shortest path and then cannot both be obtained under policy $\pi$. If $a_1\neq a_2$, then there are two different paths going from $\mathbf{T}[-1]$ to $\mathcal{N}$, containing $\mathcal{A}_1$ and $\mathcal{A}_2$ respectively. Still, they cannot both be obtained under policy $\pi$. Therefore, the claim is proved, and we can use $t(\mathcal{N})$ to denote the unique rollout step $t$ for each future node $\mathcal{N}$.

We now provide a practical implementation for computing $\boldsymbol{Q}(\mathbf{T},a)$. We first identify all the nodes $\mathcal{N}$ in the scene that satisfy the following condition: the shortest path from $\mathcal{N}$ to $\mathbf{T}[-1]$ passes through $\mathcal{A}$. We also record the rollout step $t(\mathcal{N})$ for each node as the hop of the shortest path from $\mathbf{T}[-1]$ to $\mathcal{N}$. We sort these nodes in ascending order based on the values of $t$ and sequentially compute their rollout probabilities $\text{P}_{\pi}(\mathcal{N},t(\mathcal{N})|\mathbf{T}, a)$. Finally, we use Eq (\ref{eq:Q-sup}) to obtain $\boldsymbol{Q}(\mathbf{T},a)=\sum_{\mathcal{N}}\text{P}_{\pi}(\mathcal{N},t(\mathcal{N})|\mathbf{T}, a)\gamma^{t(\mathcal{N})}\boldsymbol{R}(\mathcal{N})$. As stated in Section~\ref{sec:method-Q-train}, $\boldsymbol{R}(\mathcal{N})$ is an abstracted text-based feature. We set it to the average textual feature of the 36 views' natural language descriptions. The resulting $\boldsymbol{Q}(\mathbf{T},a)$ serves as the ground-truth Q-feature for candidate action $a$.

The Q-model is trained on the training split of MatterPort3D~\cite{MP3D,VLN}, which is also shared by REVERIE~\cite{REVERIE} and SOON~\cite{SOON}'s training set. For experiments with additional scenes, we employ the scenes, graphs, and images generated by ScaleVLN~\cite{ScaleVLN}, which consists of 800 scenes from HM3D~\cite{HM3D} and 491 scenes from Gibson~\cite{Gibson}. We do not use the trajectory annoations generated by ScaleVLN. For validation, we evaluate the Q-model on the val-unseen split of REVERIE.

\subsection{Details of the Future Encoder}
\label{sec:supp-FE}
The proposed future encoder has the same Graph Transformer architecture as DUET's global encoder, but takes different input. We build an additional graph $\tilde{G}^t$ that shares topology with DUET's navigation graph $G^t$. For each unvisited node, we maintain a similar list as described in the step (2) of Section~\ref{sec:supp-duet}, while the contents of it are the Q-features related to the node instead of the view features. For each visited node, we extract the average feature for textual descriptions (\textit{i.e.}, $\boldsymbol{R}(\mathcal{N)}$) as the node feature.

The output of GE, FE, and LE are fused together by weighted addition. Thus, the Sigmoid function employed by DUET's dynamic fusion (Section~\ref{sec:supp-duet}, step (5)) is replaced by a Softmax function.

\subsection{Training Tasks}
The training of DUET consists of two stages: offline pre-training and online finetuning. In the pre-training stage, a batch of partial trajectories are sent to the model, which is trained to perform one of the following training tasks:

\begin{itemize}
    \item MLM (masked language modeling). A random mask is applied to the instruction text, and the agent is asked to reconstruct the masked tokens. For this task, the cross-modal layers in GE/FE/LE use the node/view features as key and value, while the textual features are used as query. The output of them are summed together and processed by an MLP head for word prediction.
    \item SAP (single-step action prediction). The agent is asked to choose the best next-step action (among the global candidates) given a partial trajectory. The output action logits  are supervised by cross-entropy loss, and the ground-truth is the candidate with the shortest distance to the destination. This loss is computed on the global, local, and fused logits in DUET. We further apply it to the future logits output by FE.
    \item OG (object grounding). The agent is asked to predict the correct object given a trajectory ending at a correct location. The output object logits are supervised by cross-entropy loss.
    \item MRC (masked region classification). Similar to MLM, some of the input views and objects are masked, and the agent is asked to predict their semantic class. An MLP is appended after LE for prediction. The ground-truth semantic labels are the output class probability of a frozen classification model and a frozen detection model.
\end{itemize}

Our training stage 2 (Section ~\ref{sec:method-train}) inherits the design of these tasks. The proposed progress-related sub-tasks are integrated into SAP. To be specific, we compute the ground-truth historical progress $s_1$ and distance to go $s_2$ for each global candidate (Section~\ref{sec:method-FRM}). We then clip them to $[0,1]$, and discretize them into 5 bins. The output node features of GE and FE are sent to two separate MLPs to perform a 5-category classification task. The two cross-entropy losses are added to SAP's original loss. We expect the classification-based progress estimation to be more robust than regressing float values. Considering the range of $\text{dist}(\mathcal{S},\mathcal{C}),\text{dist}(\mathcal{C},\mathcal{A}),\text{dist}(\mathcal{A},\mathcal{G})$, the normalizing constants $D_1$ and $D_2$ are set to 2 times the length of expert trajectory, and the length of expert trajectory, respectively. 

In the finetuning stage, the agent performs sequential decision making in the scene. At each time step, the predicted action (a probability distribution on all the global candidates and ``stop") is supervised through cross-entropy loss by a pseudo expert policy, which identifies the candidate node that minimizes the sum of the distances to the current node and the destination based on the complete graph of the scene. 
The agent then finds the shortest path from its current location to its chosen candidate on the graph it builds, and traverses along it to reach the next state.
During finetuning, the agent chooses candidates by sampling from the fused action probability. While for inference, it selects the candidate with the maximum probability.

\subsection{Model Statistics}
\begin{table}[]
\caption{Distribution of parameters in the NavQ agent. The listed modules from left to right are the panoramic encoder (Section~\ref{sec:supp-duet}, step (1)), the textual encoder (Section~\ref{sec:supp-duet}, step (1)), the global encoder (Section~\ref{sec:supp-duet}, step (3)), the future encoder (Section~\ref{sec:supp-FE}), the local encoder (Section~\ref{sec:supp-duet}, step (4)), the Q-model (Section~\ref{sec:supp-Q}), and the prediction heads for generating and fusing logits.}
\vspace{-5mm}
\label{tab:stat}
\begin{center}

\scalebox{0.85}{
\begin{tabular}{ccccccc|c}
             PE & TE & GE & FE & LE & QM & Heads & Total \\ \hline
15.2 & 87.6 & 37.9 & 39.2 & 37.8 & 30.5 & 4.1 & 252.4M
\end{tabular}
}
\vspace{-5mm}
\end{center}
\end{table}
In Table~\ref{tab:stat}, we present the count of parameters for each module of our NavQ agent. Compared with DUET, the newly proposed FE and QM bring about 38\% additional parameters, while they clearly boost the overall performance as shown in Table~\ref{tab:main}.
Note that the Q-model is frozen when training the agent, reducing the impact on training cost. 
As for inference, we assess the efficiency by recording the average time for a forward pass of the full model. At each navigation step, DUET spends $\sim0.032$s to make a decision, while NavQ spends $\sim0.052$s under the same environment.

\section{Details on the Datasets and Metrics}
\subsection{Datasets}
\label{sec:supp_datasets}
Experiments are performed on two goal-oriented VLN datasets, REVERIE~\cite{REVERIE} and SOON~\cite{SOON}.
REVERIE provides high-level descriptions of the target locations and objects as instructions. We adopt the same train/val/test split strategy as DUET~\cite{DUET}. The training set consists of 60 scenes and 10,466 instructions. The unseen validation set consists of 3,521 instructions in 10 scenes with no overlap to the training scenes. The test set consists of 16 novel scenes with 6,292 instructions. The average instruction length is around 21 words, and the expert trajectory typically requires 4–7 navigation steps. Pre-defined object bounding boxes are provided for each navigable location, and the agent needs to select one box as its predicted object. During training stage 2, We incorporate the additional synthetic instructions generated by a speaker model following DUET~\cite{DUET}, which expand the training data from 10,466 to 30,102 instruction-path pairs.

SOON~\cite{SOON} is designed for a task named ``From Anywhere to Object" (FAO). It requires the agent to find the target object no matter where its starting point is. The instructions are unrelated with the agent's initial location, but only describe the position and attributes of the target object, its relation to other objects, and its residing region. Each instruction contains an average of 47 words. The corresponding paths range from 2 to 21 steps. Object bounding boxes are not provided for SOON, and the agent must predict a direction representing the target object’s center at the ending place of its trajectory. The training set of SOON comprises 3,085 instructions. Each instruction is paired with different starting points, resulting in 28,015 trajectories across 38 houses. The validation set and test set are composed of 339 instructions from 5 novel scenes, and 1,411 trajectories from 14 novel scenes. Each instruction is labeled with 10 different starting locations and 10 corresponding expert trajectories.

\subsection{Evaluation Metrics}
For navigation performance, we adopt the following standard metrics:
\begin{itemize}
\item \textbf{Success Rate (SR)}: The ratio of paths that successfully reach a correct location. For REVERIE, the correct locations are those where the target object is visible. For SOON, a ground-truth goal node is defined for each instruction by experts. The correct locations are the nodes within 3 meters of the goal node.
\item \textbf{Oracle SR (OSR)}: The SR computed under an oracle stop policy.
\item \textbf{SR Penalized by Path Length (SPL)}: The SR adjusted to account for the path length. The original 0-1 success state is weighted by $\frac{\text{length of agent's traj}}{\text{length of expert's traj}}$.
\end{itemize}

We also utilize the following metrics that take object grounding into consideration:
\begin{itemize}
\item \textbf{Remote Grounding Success (RGS)}: The proportion of instructions executed successfully. For REVERIE, it requires the agent to output the correct object instance. For SOON, it requires that the output direction falls in the range of the correct object's bounding box.
\item \textbf{RGS Penalized by Path Length (RGSPL)}: The RGS adjusted to consider the path length, similar to SPL.
\end{itemize}

\begin{table}[]
\caption{An ablation study on the effect of Q-learning techniques. Results are obtained on REVERIE's val-unseen split.}
\vspace{-5mm}
\label{tab:supp-abl}
\begin{center}
\scalebox{0.9}{
\begin{tabular}{l|ccccc}
               Q-Model & OSR & SR & SPL & RGS & RGSPL \\ \hline
w.o. & 54.42 & 48.14 & 33.38 & 30.19 & 21.05 \\
vision-based & 53.45 & 48.11 & 33.79 & 31.64 & 22.56\\
rand policy-based & 58.68 & 51.29 & 36.23 & 34.34 & 24.59\\
ours & \textbf{60.47} & \textbf{53.22} & \textbf{38.89} & \textbf{36.84} & \textbf{27.12}\\
\end{tabular}
}
\vspace{-8mm}
\end{center}
\end{table}

\section{Additional Experimental Results}

\subsection{Ablation Study on the Training of Q-Model}
Here we give an analysis on the various techniques proposed in Section~\ref{sec:method-Q} for pre-training our Q-model. In Section~\ref{sec:method-Q-train}, two designs are put forward for enhancing the generalizability of the Q-features. We have visualize the effect of the MAE pre-training by showing the loss curve in Figure~\ref{fig:Qloss}, while the benefits of text-based prediction cannot be easily seen from the MSE loss, since the visual features and textual features have different scale. Thus, we compare using a visual prediction-based Q-model (\textit{i.e.}, $\boldsymbol{R}$ set as the aggregated average view features) against our default setting. As is Table~\ref{tab:supp-abl}, employing textual features has a clear advantage over the vision-based Q-model. 

Besides, we try out using a random policy instead of the $\pi$ described in Section~\ref{sec:method-Q-data} and Section~\ref{sec:supp-Q}. For each state-action pair, we use simulations to approximate the expectation in Eq (\ref{eq:Q}), where the agent uniformly chooses a local candidate at each rollout step. As in Table~\ref{tab:supp-abl}, integrating this random policy-based Q-model will lead to higher navigation performance than the baseline without Q-model, but the gain is less significant than our default setting. Therefore, the preference for optimal paths in $\pi$ is indeed helpful for executing goal-oriented VLN tasks.

\subsection{Results with Other Backbones}
NavQ is a modular model enhancement that can be integrated with any baseline method focusing on leveraging historical information. In the main text, we mainly adopt DUET~\cite{DUET} as the baseline. In accordance with the reviewers’ suggestions, here we explore an alternative backbone, BEVBert~\cite{BEVBert}, which models the local environment with a top-down metric map, complementing the global topological representation. As shown in Table~\ref{tab:supp-bev}, incorporating the Q model and the future branch into BEVBert leads to notable improvements on most evaluation metrics, demonstrating the generalizability of the proposed method to some extent.

\begin{table}[]
\caption{The results on REVERIE with BEVBert as backbone.}
\vspace{-5mm}
\label{tab:supp-bev}
\begin{center}
\scalebox{0.85}{
\begin{tabular}{c|l|ccccc}
               & & OSR & SR & SPL & RGS & RGSPL \\ \hline
Val & BEVBert~\cite{BEVBert} & 56.40 & 51.78 & 36.37 & 34.71 & 24.44\\
Unseen & NavQ (Ours) & \textbf{60.07} & \textbf{54.08} & \textbf{38.49}& \textbf{35.36}& \textbf{25.45}

\\ \hline
Test & BEVBert~\cite{BEVBert} & 57.26 & \textbf{52.81} & \textbf{36.41} & 32.06 & 22.09\\
Unseen & NavQ (Ours) & \textbf{60.04} & 52.42 & 36.40 & \textbf{36.59} & \textbf{24.95}
\end{tabular}
}
\vspace{-5mm}
\end{center}
\end{table}

\subsection{Results on Other Dataset}
Based on the reviewers’ suggestions, here we discuss the potential of NavQ on other VLN datasets.
Apart from REVERIE~\cite{REVERIE} and SOON~\cite{SOON}, there are some classical datasets, such as R2R~\cite{VLN} and RxR~\cite{RxR}, in which the instructions are procedure-based rather than goal-based. As a result, the agent is required to follow the route described in the instructions, rather than merely reaching a specified destination.
We note that our proposed method is tailored for \emph{goal-oriented} VLN, and the formulation of Q-learning encourages the agent to reach the destination as quick as possible. Thus, NavQ is not quite suitable for procedure-based benchmarks, especially RxR, since it features non-shortest expert paths.
We conduct preliminary experiments with NavQ on the R2R dataset, as it still satisfies the shortest-path assumption. As shown in in Table~\ref{tab:supp-r2r}, NavQ achieves better performance than the base model, especially on the efficiency-related metric. However, the improvement is not as significant as on REVERIE, since the goal-centric future branch may not fully utilize the process-related information in the instructions.

\begin{table}[]
\caption{The results on R2R.}
\vspace{-5mm}
\label{tab:supp-r2r}
\begin{center}

\scalebox{0.9}{
\begin{tabular}{c|l|cccc}
               & & TL & NE$\downarrow$ & SR$\uparrow$ & SPL$\uparrow$ \\ \hline
Val & DUET~\cite{DUET} & 13.94 & 3.31 & 72 & 60 \\
Unseen & NavQ & 13.80 & \textbf{3.06} & \textbf{73} & \textbf{63} \\ \hline
Test & DUET~\cite{DUET} & 14.73 & 3.65 & 69 & 59 \\
Unseen & NavQ & 14.41 & \textbf{3.30} & \textbf{72} & \textbf{63}
\end{tabular}
}
\vspace{-5mm}
\end{center}
\end{table}

\subsection{More Visualization Results}

\begin{figure}[t]
\begin{center}
   \includegraphics[width=\linewidth]{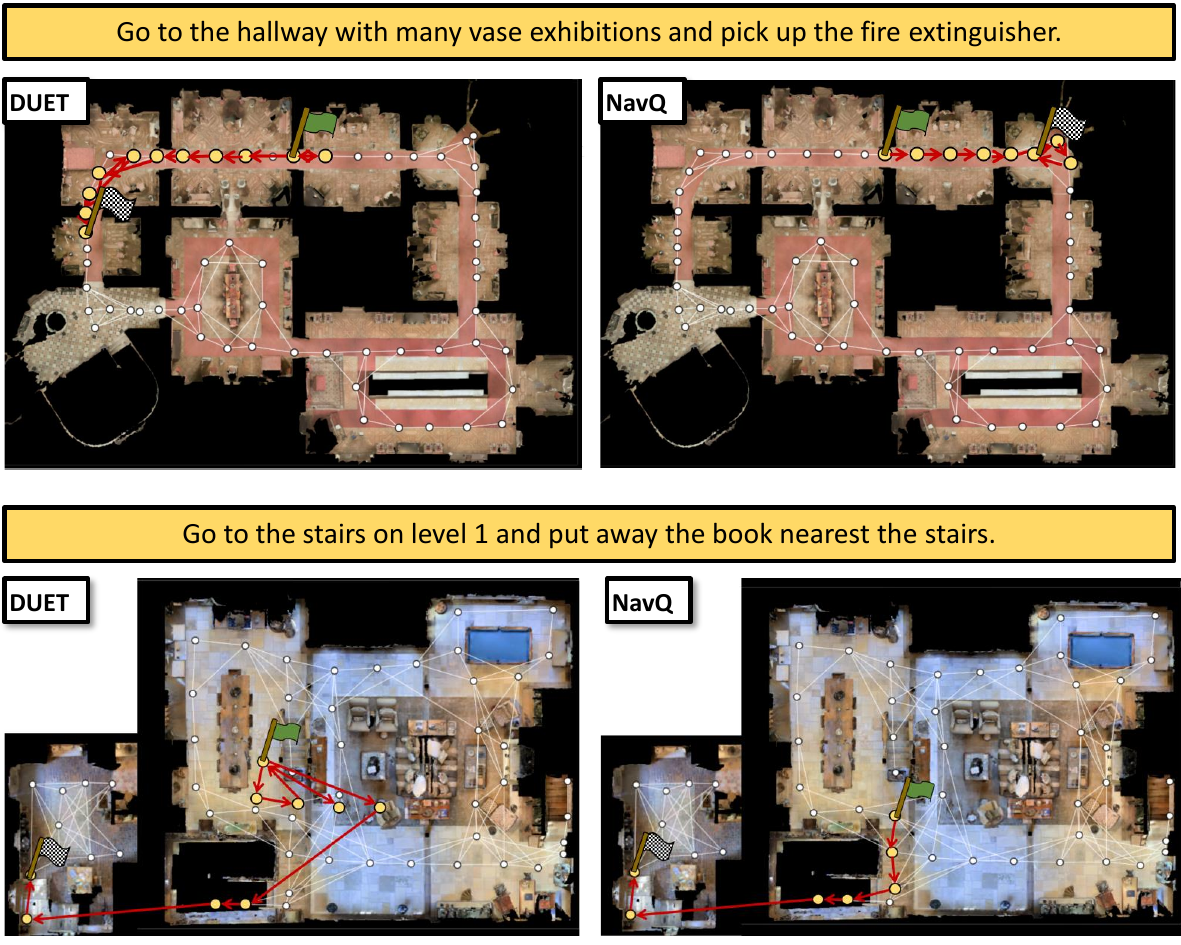}
\end{center}
\vspace{-5mm}
\caption{A qualitative comparison of our method and the baseline agent. In the upper example, NavQ reaches the correct destination while the baseline does not. In the lower example, NavQ arrives at the target object with less steps than DUET.}
\label{fig:supp-qual}
\end{figure}
\begin{figure}[t]
\begin{center}
   \includegraphics[width=0.9\linewidth]{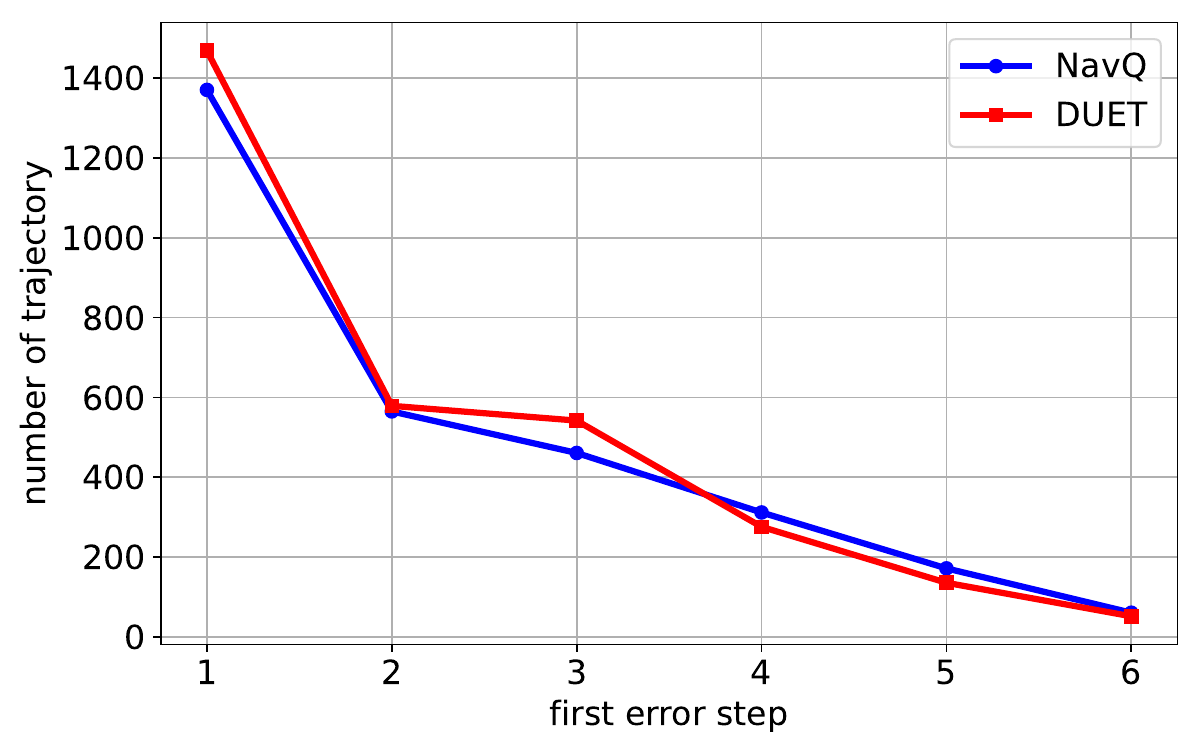}
\end{center}
\vspace{-5mm}
\caption{The distribution of navigation errors on REVERIE's val-unseen set.}
\label{fig:error-stat}
\end{figure}

In Figure~\ref{fig:supp-qual}, we provide two more qualitative comparisons between the NavQ agent and the baseline agent.

In addition, we discuss the distribution of navigation errors.
Among the 3,521 validation instructions, our model produces 580 predicted trajectories that were identical to the expert trajectories, whereas DUET produces 468. For the remaining 2,941/3,053 trajectories, we analyze the position where the model makes the first error, \textit{i.e.}, deviates from the expert trajectory. The results are presented in Figure~\ref{fig:error-stat}. It can be noticed that our model makes fewer mistakes at the beginning and middle stage of the episode. This aligns well with the motivation of our foresighted agent, which is to make better decisions when the historical information (observations up to now) is not sufficient enough.

\end{document}